\pdfoutput=1

\documentclass[11pt]{article}

\usepackage[final]{acl}

\usepackage{wrapfig,lipsum,graphicx}
\usepackage{amsmath}
\usepackage{amssymb}
\usepackage{multirow}
\usepackage{graphicx}
\usepackage{booktabs}
\usepackage[normalem]{ulem}
\useunder{\uline}{\ul}{}
\usepackage{hyperref}

\usepackage{times}
\usepackage{latexsym}

\usepackage[T1]{fontenc}

\usepackage[utf8]{inputenc}

\usepackage{microtype}

\title{Learning Job Title Representation from Job Description Aggregation Network}

\author{Napat Laosaengpha\textsuperscript{$\spadesuit$}, Thanit Tativannarat\textsuperscript{$\spadesuit$}, Chawan Piansaddhayanon\textsuperscript{$\heartsuit$} \\ {\bf Attapol Rutherford}\textsuperscript{$\diamondsuit$},\and {\bf Ekapol Chuangsuwanich}\textsuperscript{$\spadesuit$, $\heartsuit$} \\ 
\textsuperscript{$\spadesuit$}Department of Computer Engineering, Faculty of Engineering, Chulalongkorn University \\
\textsuperscript{$\heartsuit$}Center of Excellence in Computational Molecular Biology, \\ Faculty of Medicine, Chulalongkorn University\\
 \textsuperscript{$\diamondsuit$}Department of Linguistics, Faculty of Arts,
Chulalongkorn University,
Thailand \\
 \small{\texttt{
napatnicky@gmail.com \quad{thanit.tati@gmail.com} \quad{schwanph@gmail.com}}
 }
 \\
 \texttt{\small{{attapol.t@chula.ac.th} \quad{ekapolc@cp.eng.chula.ac.th}}}
 \\
}

\begin{document}
\maketitle
\begin{abstract}

Learning job title representation is a vital process for developing automatic human resource tools. To do so, existing methods primarily rely on learning the title representation through skills extracted from the job description, neglecting the rich and diverse content within. Thus, we propose an alternative framework for learning job titles through their respective job description (JD) and utilize a Job Description Aggregator component to handle the lengthy description and bidirectional contrastive loss to account for the bidirectional relationship between the job title and its description.
We evaluated the performance of our method on both in-domain and out-of-domain settings, achieving a superior performance over the skill-based approach.
\end{abstract}

\section{Introduction}

With the rapid expansion of online recruitment platforms, vast amounts of job advertisement data (JAD) have been generated. One key part of this data is a job posting, providing detailed information on job titles, specialties, and responsibilities for open positions. Thus, the availability of a system that could understand the post semantics, especially job titles, would greatly facilitate the matchmaking process between both the recruiters and job applicants. This leads to a surge of interest in learning job title representation due to its potential ability to automate job-related tasks such as job recommendation \citep{Kaya2021EffectivenessOJ, Zhao2021EmbeddingbasedRS}, job trajectory prediction \citep{Decorte2023CareerPP}, and job title benchmarking \citep{Bana2021job2vecLA}.

To learn the title representation, previous works have primarily relied on utilizing skills information to learn the association between the job title and their respective skill \citep{Decorte2021JobBERTUJ, Zbib2022LearningJT, Bocharova2023VacancySBERTTA}. However, this approach also has some shortcomings as it requires skill information. The skills for a given job are either manually listed, which can be erroneous or incomplete, or automatically extracted from the job description through methods such as keyword matching or automatic skill extraction \citep{Zhang2022SkillEF, Li2023SkillGPTAR}. These skill extraction methods often require a predefined skill vocabulary or a curated dataset \citep{Zhang2022SkillSpanHA}. Furthermore, it is necessary to keep these resources up-to-date with trends in the job market as the dynamic and rapid growth of emerging job roles.

Previous works mitigate these problems by generating synthetic skill data \citep{Decorte2023ExtremeMS, Clavi2023LargeLM} or creating datasets where both job titles, and skill lists are readily available \citep{Bhola2020RetrievingSF, Goyal2023JobXMLCEM}. Nonetheless, the former approach further increases pipeline complexity, while the latter suffers from missing skills annotation caused by a communication gap between employers and recruiters.

In this work, we propose to overcome the challenges of obtaining a comprehensive set of skills by bypassing the whole process and instead develop a new framework to learn job titles directly through job descriptions (JDs) without the need for the skill extraction pipeline.  We introduce job description aggregation network, which reweights each segment of the JD by their importance and then aggregates them into a unified JD representation. Our contributions are as follows:

$\bullet$ Our JD-based method outperforms all previous skill-based approaches in both in-domain and out-of-domain settings, achieving up to a 1.8\% and 1.0\% absolute performance gain.

$\bullet$ Our ablation study shows that the ability to reweight segments according to their importance afforded by our model is critical to its accuracy. 

$\bullet$ We show that our approach can implicitly learn the information about the underlying skills associated with job titles.





\begin{figure*}[t]
\centering
\includegraphics[width=0.85\textwidth]{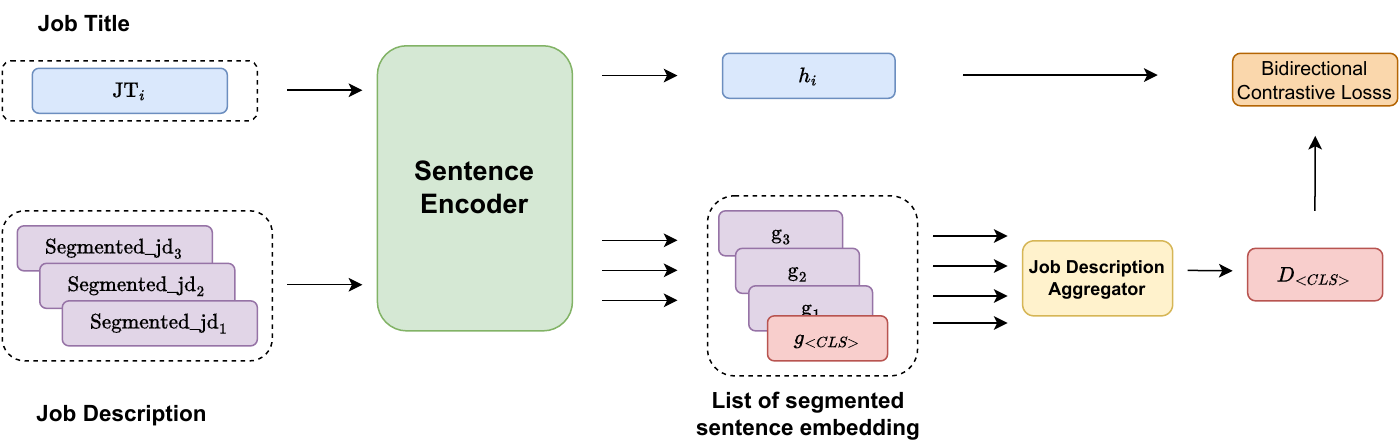}
\caption[Proposed architecture]{The overall proposed method, JD Aggregation Network (JDAN), presents a dual-encoder architecture coupled with a job description aggregator module.} 
\label{fig:architecture}
\end{figure*}

\section{Our Proposed Method}

An overview of the proposed framework is illustrated in Figure \ref{fig:architecture}. Initially, job titles and their respective segmented job descriptions are independently fed into a sentence encoder to obtain their representation. Subsequently, the sentence embeddings are fused through the job description aggregator to acquire a unified representation. Finally, bidirectional contrastive loss is utilized as a training objective to maximize a pairwise similarity between the embedded job title and their respective aggregated job description representation while minimizing others.

The subsections describe the sentence encoder, job description aggregator, and contrastive learning process in detail.

 \subsection{Sentence Encoder}

The sentence encoder follows a dual-encoder architecture \citep{guo-etal-2018-effective,Feng2020LanguageagnosticBS}, in which the job title and description are fed into the encoder separately to generate their respective representations. The job title is fed into the model to obtain a job title embedding $h$. On the other hand, the job description can be lengthy with some irrelevant parts. Thus, the job description is broken down into sentences and encoded sentence by sentence, resulting in a list of segmented sentence embeddings $G = [g_1, g_2, \dots, g_n]$. Then, the embeddings are aggregated into a final representation for the description. The sentence segmentation process is explained in Appendix \ref{apd:preprocess}.

 \subsection{Job Description Aggregation Network (JDAN)}

The job description aggregator is responsible for combining multiple sentence embeddings into a unified representation by weighting the importance of each sentence. This step is important because job descriptions often contain information not directly related to the corresponding job titles such as location and salary. Inspired by \citep{wu2021representing, Zhang2023ModelingSS}, we create an additional learnable token $g_{<CLS>}$ to represent the summarized token and prepend it to the sequence of sentence embedding $G = [g_{<CLS>}, g_1, g_2, \dots, g_n]$. A Layer Normalization and a shallow transformer encoder are then applied to $D$ to obtain a list of learned representation $D = [d_{<CLS>}, d_1, d_2, \dots, d_n]$.


\begin{equation} 
    D = \text{TransformerEncoder}(\text{LayerNorm}(G))
\end{equation}



The learned summarized token $d_{<CLS>}$ is then fed through three MLP layers with ReLU activation to obtain a final unified representation $f$.



\subsection{Bidirectional Contrastive Learning}

The bidirectional contrastive learning minimizes the following training objective function:

\begin{equation} 
    \resizebox{0.98\hsize}{!}{$
    \mathcal{L}_i= -(\log \frac{e^{sim(h_i,f_i)/\tau}}{\sum^N_{j=1}e^{sim(h_i,f_j)/\tau}}  + \log \frac{e^{sim(h_i,f_i)/\tau}}{\sum^N_{j=1}e^{sim(h_j,f_i)/\tau}})$}
\end{equation}


where $h_i$ is the $i^{th}$ job title embedding, $f_i$ is $i^{th}$ the unified job description embedding, $\tau$ is the temperature scaling parameter, and $N$ is the batch size. Instead of only maximizing the similarity between the job title embedding $h_i$ and its respective job description embedding $f_i$ while minimizing the similarities of other pairs $sim(h_i, f_j)$, the objective function also further introduces the second term to minimize the similarity of the job description embedding $f_i$ to other job titles $sim(h_j, f_i)$. This could be seen as an extension of SimCSE \citep{Gao2021SimCSESC} where an additional latter term is included to mitigate the overlooked characteristic of JAD where a bidirectional relationship (job title to job description and vice versa) exists.  \citep{Yang2019ImprovingMS,Feng2020LanguageagnosticBS}.

\section{Experimental Setup}


We benchmarked the performance of our proposed framework under two settings: \textbf{in-domain} and \textbf{out-of-domain}. 

For the in-domain evaluation, we used our own \textbf{JTG-Jobposting} dataset for training and \textbf{JTG-Synonym} for evaluation. JTG-Jobposting is a private Thai-English job posting dataset consisting of 28,844 job postings from, \url{https://jobtopgun.com}, a renowned recruitment website in Thailand. The postings include job titles, job descriptions, and skills. We performed benchmarking on the JTG-Synonym dataset by posing the problem as a cross-lingual synonym retrieval task where the job titles were used as queries, and all synonyms were used as the candidate pool. Each query was performed on the English and Thai candidate pools separately to calculate the R@5, R@10, and mAP@25. The final metric values were obtained by averaging across every query-candidate-pool pairs. This evaluation protocol was intentionally designed to avoid language bias where a query would prefer a candidate from the same language. This issue will be later discussed in the Section \ref{4.5}.



For the out-of-domain evaluation, we used \textbf{Mycareersfuture.sg} \citep{Bhola2020RetrievingSF}, a dataset of real-world job postings collected by the Singaporean government as the training set. ESCO \citep{Vrang2014ESCOBJ}, a standardized system of the European Union (EU) for and categorizing skills, competencies, qualifications and occupations was used as the validation and test data. The task chosen for benchmarking is job normalization where the goal is to predict the standardized version of each job title. The ESCO job normalization dataset consists of 30,926 unique job titles and 2,675 standardized ESCO occupation labels. We followed \citep{Decorte2021JobBERTUJ} and used the standard micro-average of recall at 5, 10 and MRR as metrics. 

The summary statistics of the datasets are shown in Table \ref{tab:data-stats}. See Appendix~\ref{sec:datasetdesc} for more details.

\subsection{Implementation Details}

\begin{table*}[t]
\centering
\LARGE
\resizebox{\textwidth}{!}{%
\scalebox{1}{
\begin{tabular}{llllllll}
\hline
\toprule
\multicolumn{2}{c}{\multirow{2}{*}{\textbf{Method}}}                    & \multicolumn{3}{c}{\textbf{JTG-Synonym}}                                           & \multicolumn{3}{c}{\textbf{Job Normalization}}                                 \\ \cline{3-8} 
\multicolumn{2}{c}{} & \textbf{R@5} $\uparrow$            & \textbf{R@10} $\uparrow$           & \textbf{mAP@25} $\uparrow$         & \textbf{R@5} $\uparrow$            & \textbf{R@10} $\uparrow$           & \textbf{MRR} $\uparrow$            \\ \hline
\multicolumn{2}{l}{XLM-R / BERT (w/o finetuning)}                      & 15.11                    & 18.68                    & 10.27                    & 26.23                    & 32.10                    & 20.57                    \\ \hline
\multicolumn{2}{l}{\textbf{Skill-based method}}                         &                          &                          &                          &                          &                          &                          \\ \hline
                     & JobBERT \citep{Decorte2021JobBERTUJ}             & 31.04 ($\pm$0.61)                    & 41.99 ($\pm$0.86)                   & 22.83 ($\pm$0.42)                   & $\text{38.65}^{\dag}$          & $\text{46.04}^{\dag}$          & $\text{30.92}^{\dag}$          \\
                     & Doc2VecSkill \citep{Zbib2022LearningJT}                       & 25.64 ($\pm$0.35)                   & 34.44 ($\pm$0.61)                   & 18.50 ($\pm$0.28)                   & $\text{45.95}^{\dag}$         & $\text{54.00}^{\dag}$             & $\text{34.14}^{\dag}$             \\
                     & VacancySBERT \citep{Bocharova2023VacancySBERTTA} & \multicolumn{1}{c}{-}    & \multicolumn{1}{c}{-}    & \multicolumn{1}{c}{-}    & $\text{42.50}^{\dag}$                & $\text{55.60}^{\dag}$                & \multicolumn{1}{c}{-}    \\
                     & Keyword Skill (ours)                              & {\ul 47.79} ($\pm$0.32)          & {\ul 60.62} ($\pm$0.32)          & {\ul 35.87} ($\pm$0.13)          & {\ul 48.23} ($\pm$0.19)          & {\ul 56.45} ($\pm$0.18)          & {\ul 37.68} ($\pm$0.15)          \\ \hline
\multicolumn{2}{l}{\textbf{JD-based method}}                            &                          &                          &                          &                          &                          &                          \\ \hline
                     & JD Aggregation Network (ours)                           & $\textbf{49.21}^*$ ($\pm$0.55) & $\textbf{62.34}^*$ ($\pm$0.31) & $\textbf{37.09}^*$ ($\pm$0.45) & $\textbf{49.24}^*$ ($\pm$0.43) & $\textbf{57.22}^*$ ($\pm$0.43) & $\textbf{38.71}^*$ ($\pm$0.24) \\ \bottomrule
\end{tabular}%
}
}
\caption{The performance of our proposed JD-based method against other skill-based methods.  The best results are bolded, and the second-best ones are underlined. $\dag$: results from their original papers. * denotes significant improvement over Keyword Skill (ours) using two-sample t-test (p < 0.01).}
\label{tab:main-result}
\end{table*}



We compared the performance of our proposed framework to other previously proposed skill-based methods, which are \textbf{JobBERT} \citep{Decorte2021JobBERTUJ}, \textbf{Doc2VecSkill} \citep{Zbib2022LearningJT}, \textbf{VacancySBERT} \citep{Bocharova2023VacancySBERTTA}, and our own skill-based approach. In our approach, we used a dual-encoder to independently encode the job title and the concatenated set of skills corresponding to the title using a specified separator token ("[SEP]" for BERT \citep{Devlin2019BERTPO} and  "</s>" for XLM-R  \citep{Conneau2019UnsupervisedCR}). Then, the bidirectional contrastive loss was used to encourage the job title and the respective encoded skills pair to come together while pushing the others away. All skill-based models used keyword matching to extract skills from the job description. 

For automatic skill extraction, we used SkillSpan\footnote{\url{https://huggingface.co/jjzha/jobbert_skill_extraction}} to extract keywords from the Mycarrersfuture.sg dataset. On the other hand, as the JTG-job posting dataset contains both English and Thai, we made some modifications by creating a new classifier to extract the skills from the job description by posing the problem as a multi-label classification instead. Additional details for the automatic skill extraction is provided in Appendix \ref{apd:skillextraction}. 



The experiments were conducted using the pretrained-language model BERT as a sentence encoder \citep{Devlin2019BERTPO} on the job normalization task and XLM-R \citep{Conneau2019UnsupervisedCR} on the synonym retrieval task because JTG-Jobposting contains both Thai and English. The sentence representations of every model were obtained through mean pooling of every token (words) in the sentence. We reported the average with standard deviation from five random seeds. The hyperparameters were obtained through grid search in the validation set. The extended implementation detail of our method is shown in the Appendix \ref{apd:tuning}, \ref{apd:impdetail}.

\section{Results and Analysis}




\subsection{Main Results}
Our method achieved consistent performance improvement over all previous skill-based approaches on both datasets under all metrics, achieving
up to 1.8\% and 1.0\% absolute performance gain on the JTG-Synonym and Job normalization task, respectively (Table \ref{tab:main-result}).

\subsection{Comparison between Skill-based and JD-based method}

We found that our framework also outperformed or achieved competitive performance when compared to other skill-based approaches, even with human annotations (Table \ref{tab:compare}). Surprisingly, our method performed better in skill-based recruiter annotation on the job normalization task but not the JTG-Synonym task. This is because skill annotations for the MyCareersFuture.sg dataset might be incomplete due to a communication gap between employers and recruiters \citep{Bhola2020RetrievingSF}. In addition, some parts of the JTG-Jobposting dataset also contain implicit information that is only present in the recruiter's annotated skill and not explicitly mentioned in the JD. For example, a job posting for "Data analyst" is annotated with the skills "Excel", "SQL", and "Python," but these skills do not appear in any part of the JD. Additional examples could be seen in Figure \ref{fig:compare-skill} of our Appendix.

\begin{table}[ht]
\centering
\resizebox{\columnwidth}{!}{%
\begin{tabular}{@{}lcc@{}}
\toprule
\multirow{2}{*}{}    & \textbf{JTG-Synonym}              & \textbf{Job Normalization}        \\ \cmidrule(l){2-3} 
                     & \textbf{R@10 $\uparrow$}      & \textbf{R@10 $\uparrow$}             \\ \midrule
\textbf{Skill-based} & \multicolumn{1}{l}{} & \multicolumn{1}{l}{}     \\
- Recruiter Annotation          & \textbf{63.06} ($\pm$0.67)                & 51.51 ($\pm$0.30
)          \\
- Keyword Matching   & 60.62 ($\pm$0.32)                & {\ul 56.45} ($\pm$0.18)          \\
- Skill Extraction   & 51.84 ($\pm$0.87)                    & 55.11 ($\pm$0.27)          \\ \midrule
\textbf{JD-based (ours)}    & {\ul 62.34} ($\pm$0.31)                & \textbf{57.22} ($\pm$0.43) \\ \bottomrule
\end{tabular}%
}
\caption{Comparison between learning job titles through skills or job descriptions.}
\label{tab:compare}
\end{table}

\subsection{Probe Analysis on Job Title Embeddings}
\begin{figure}[ht]
\centering
\includegraphics[scale=0.28, height=0.30\textwidth]{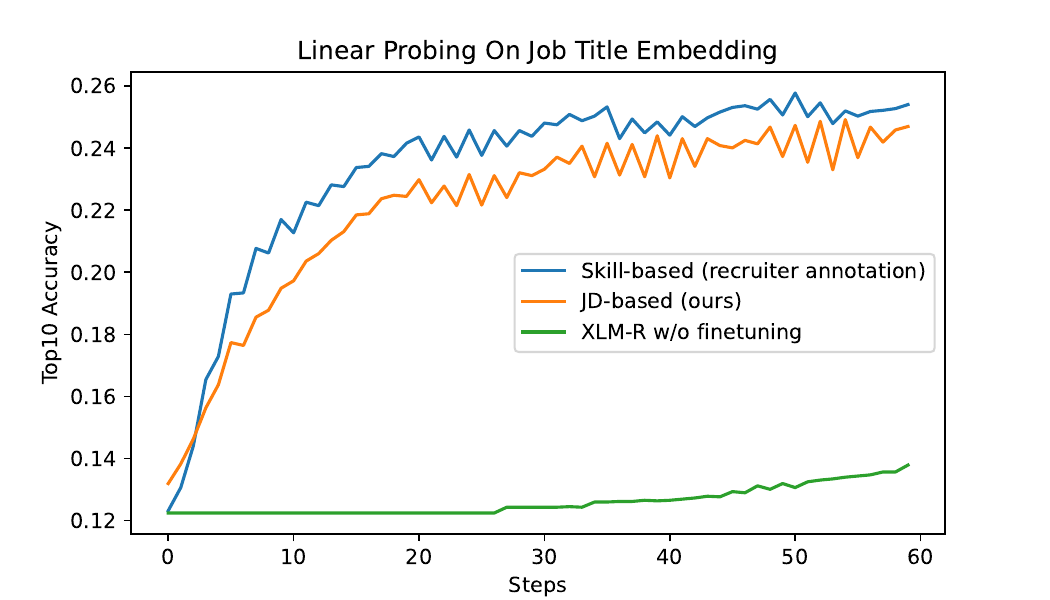}
\caption[Proposed architecture]{Comparison of job title linear probing from the model learning from the skill-based method, our JD-based method, and without any fine-tuning. }
\label{fig:probe}
\end{figure}

We also conducted an additional analysis on the learned embeddings by linear probing \citep{DBLP:conf/iclr/AlainB17} (training a linear classifier on top of the embeddings) . We trained classifiers to predict skills from the job title embeddings learned from various methods. As shown in Figure \ref{fig:probe}, the result of linear probing on a held-out set of the JTG-Jobposting dataset (details provided in Appendix \ref{apd:linprobe}) found that the model learned through recruiter-annotated skills and JD were on par (25.8 vs 24.5 Top-10 accuracy) which doubled the performance of using XLM-R embedding without finetuning (13.1 Top-10 accuracy), implying that the JD-based method could implicitly understand skill information despite not being trained with one. This offers an explanation on why learning the job title representation from the job description can be more beneficial than from skills as the skill information can be learned implicitly while having access to other additional information.

\begin{table*}[t]
\centering
\scalebox{0.90}{
\begin{tabular}{c|c|c}
\hline
\toprule
\multicolumn{1}{l|}{}                               & \textbf{JTG-Jobposting}    & \textbf{Mycarreersfuture.sg} \\ \hline
\# of job postings                                      & 28,844          & 20,298                       \\ \hline
\# of total skills                    &     301,124      &            405,606             \\ \hline
\# of distinct skill                    & 35,107          & 2,548                        \\ \hline
\# of skills with more than 20 occurrences                 & 3,092          & 1,209                        \\ \hline
Average skill tags per job posting                     & 10.43 $\pm$ 19.91         & 19.98 $\pm$ 0.06                      \\ \hline
Average token count of job description per job posting & 123.10 $\pm$ 82.13         & 155.77 $\pm$ 103.55                      \\ \hline
Average \# of segmented sentence per job posting        & 5.88 $\pm$ 3.87 & 6.63 $\pm$ 27.14             \\ \hline
Average token count of segmented sentence           & 21.23 $\pm$ 26.58          & 27.19 $\pm$ 27.14                       \\ \bottomrule
\end{tabular}%
}
\caption{Dataset statistics of JTG-Jobposting and Mycarreersfuture.sg.}
\label{tab:data-stats}
\end{table*}

\begin{table*}[]
\centering
\footnotesize
\scalebox{1.15}{

\begin{tabular}{lrrrrrr|rrr}
\hline
\toprule
\textbf{Candidate pool} & \multicolumn{3}{c}{\textbf{Thai pool}}           & \multicolumn{3}{c|}{\textbf{English pool}}       & \multicolumn{3}{c}{\textbf{Combined pool}}       \\ \hline
\textbf{Query}          & \textbf{EN}    & \textbf{TH}    & \textbf{CS}    & \textbf{EN}    & \textbf{TH}    & \textbf{CS}    & \textbf{EN}    & \textbf{TH}    & \textbf{CS}    \\ \hline
XLM-R                   & 2.56           & 34.84          & 29.58          & 31.46          & 3.05           & 8.24           & 17.00          & 19.73          & 15.35          \\
JobBERT                 & 29.29          & 57.11          & 50.85          & 49.59          & 27.72          & 48.53          & 30.78          & 32.44          & 33.44          \\
Skill-based (ours)      & 56.14          & 68.05          & 62.35          & \textbf{64.22} & 52.59          & 59.13          & 35.35          & 41.57          & 38.24          \\
JD-based (ours)         & \textbf{59.87} & \textbf{71.15} & \textbf{72.35} & 64.12          & \textbf{56.46} & \textbf{70.40} & \textbf{37.93} & \textbf{42.80} & \textbf{40.92} \\ \bottomrule
\end{tabular}%
}
\caption{The comparison of cross-lingual performance (R@10 $\uparrow$) on JTG-Synonym retrieval task. "CS" refers to queries which contain Thai and English code-switching.}
\label{tab:multi-result}
\end{table*}

\subsection{Aggregation Design Choices} 
Next, we analyze our design of the job description aggregator and compare it against three other possibilities (Table \ref{tab:doc_level}). Instead of using transformers to aggregate multiple sentence embeddings, we can use mean or max pooling. Another possibility is to use the encoder to encode the entire job description (Document Level). Our proposed method outperforms the others, highlighting the importance of having a segmented sentence representation and weighting mechanism for each sentence. We provide examples of how the first attention layer selects important parts of the job description in Figure \ref{fig:heatmap} and Section \ref{sec:attention_map} of our Appendix.

\begin{table}[ht]
\centering
\resizebox{\columnwidth}{!}{%
\begin{tabular}{@{}llc@{}}
\toprule
\multirow{2}{*}{}       & \multicolumn{1}{c}{\textbf{JTG-Synonym}} & \textbf{Job Normalization} \\ \cmidrule(l){2-3} 
                        & \multicolumn{1}{c}{\textbf{R@10 $\uparrow$}}           & \textbf{R@10 $\uparrow$}                        \\ \midrule
\textbf{Document Level} & 61.44 ($\pm$0.51)                      & {\ul 57.09} ($\pm$0.39)            \\ \midrule
\textbf{Sentence Level} &                                      & \multicolumn{1}{l}{}       \\
- max pooling     & {\ul 62.18} ($\pm$0.30)                      & 55.94 ($\pm$0.29)      \\
- mean pooling          &  61.42 ($\pm$0.36)                                    & 56.66 ($\pm$0.41)     \\
- JD aggregator (ours)           & \textbf{62.34} ($\pm$0.31)                      & \textbf{57.22} ($\pm$0.43)            \\ \bottomrule
\end{tabular}%
}
\caption{A Comparison of our proposed method (JD aggregator) against the other three baselines.}
\label{tab:doc_level}
\end{table}

\subsection{Cross-lingual Evaluation}
\label{4.5}
Studies have pointed out the problem of language bias in textual embeddings \cite{roy-etal-2020-lareqa,yang-etal-2021-universal}. Embeddings from the same language are generally closer together compared to their cross-lingual counterparts. As a result, a query in Thai will prefer Thai candidates over English ones, and vice versa. To avoid this bias, the candidate pool was divided into Thai and English pool, and retrieval was done separately. Table \ref{tab:multi-result} shows the retrieval results for different query-candidate-pool pairs. As expected, all models perform better despite the pool being split. Cross-lingual setups are generally more challenging. However, our method consistently outperforms others in every setting. 

\newpage

\section{Conclusion}
\label{sec:bibtex}
In this paper, we propose a framework for learning the semantic similarity of job titles through job descriptions, bypassing the need for a complete set of skills. The job description aggregator and bidirectional contrastive loss are also introduced to handle the nature of lengthy job descriptions and the two-way relationship between the job title and its description. Our results show that our approach achieves superior performance over the previous state-of-the-art skill-based methods.


\section{Limitations}


The limitations of our work are as follows:

$\bullet$  Our framework is limited to information only available in the job description. Thus, in some cases, our performance might be sub-optimal than recruiter annotations, which could provide information that is not explicitly mentioned in the job description. 

$\bullet$   Our job description aggregator requires the description to be segmented. This could be challenging when applied to languages other than English.

$\bullet$ The job description aggregator is not designed for encoding the entire job description, which does not guarantee that our job description aggregator can be further used in downstream tasks that require job description embedding.


\section{Ethics Statement}

An inclusion of the job description for learning the job title could induce gender and age bias into job recommendations and search results. This might affect the fairness and inclusiveness of the job matchmaking process. \citep{Saxena2021ExploringAM,Seyedsalehi2022BiasawareFN}.


\section{Acknowledgements}
This work is supported in part by JOBTOPGUN, job postings and recruitment platform in Thailand. We also would like to thank the Chulalongkorn Computational Molecular Biology Group (CMB@CU) for providing additional computational resources.




\bibliography{custom}
\appendix


\section{Data Pre-processing}
\label{apd:preprocess}

For the JTG-Jobposting dataset, we used the field "Job\_Description" as a job description, and a heuristic algorithm was then applied to segment it by splitting them based on bullet points, hyphens, and numbering using regular expression.

For the MyCareersFuture.sg dataset, we concatenated the fields "Role \& Responsibilities" and "Job Requirement" to represent the job description as suggested in their work. However, since stop words and punctuations have been removed from this data, we used a punctuation restoration model\footnote{\url{https://huggingface.co/felflare/bert-restore-punctuation}} and then apply NLTK sentence segmentation for segmenting the job description.

\section{Hyperameters Tuning}
\label{apd:tuning}

Table \ref{tab:hyper} shows the hyperparameters chosen for grid search. The search was based on the validation performance (mAP@25 or MRR). The AdamW optimizer was used with a linear warm-up for 10\% of the training steps with a batch size of 16. Every model was trained for ten epochs on the JTG-Synonym task and five epochs on the job normalization task. We calculated R@5, R@10, MRR, and mAP@25 using the trec\_eval Python package \citep{VanGysel2018pytreceval}. Every experiment was conducted using PyTorch \citep{paszke2019pytorch} and NVIDIA RTX 3090GPU with 24GB memory.

All models ended up with a temperature parameter, $\tau$, of 0.05. For our skill-based approach, an initial learning rate of 3e-5 and 1e-5 was used in the JTG-Synonym and job normalization task, respectively. For our JD-based approach, an initial learning rate of 1e-5 and 3e-5 were used in the JTG-Synonym and job normalization task, respectively. The number of transformer layers used in the job description aggregator was 4.





\begin{table}[ht]
\centering
\resizebox{\columnwidth}{!}{%
\begin{tabular}{l|c}
\hline
              & List of values                             \\ \hline
Learning Rate & [1e-4, 1e-5, 3e-5, 5e-5, 3e-6, 5e-6] \\ \hline
Temperature ($\tau$)  & [0.1, 0.05, 0.01]                          \\ \hline
\# layers of JD aggregator  & [1, 2, 4]                           \\ \hline
\end{tabular}%
}
\caption{Hyperparameters and their values used in grid search.}
\label{tab:hyper}
\end{table}


\begin{figure*}[t]
    \centering
    \includegraphics[width=1.10\textwidth]{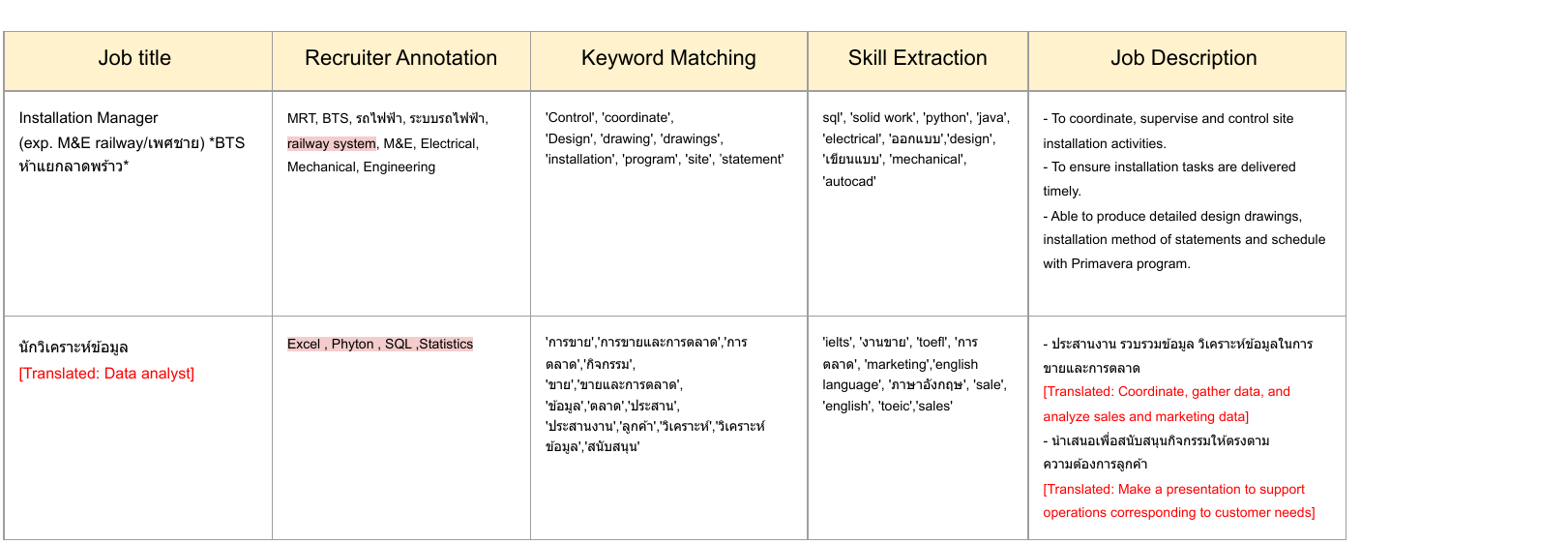} 
    \caption{Examples of annotated skills that are not explicitly mentioned in the job description but are presented in the recruiter annotation (red highlights).}
    \label{fig:compare-skill}
    
\end{figure*}

\begin{figure*}[t]
    \centering
    \fbox{\includegraphics[width=0.9\textwidth]{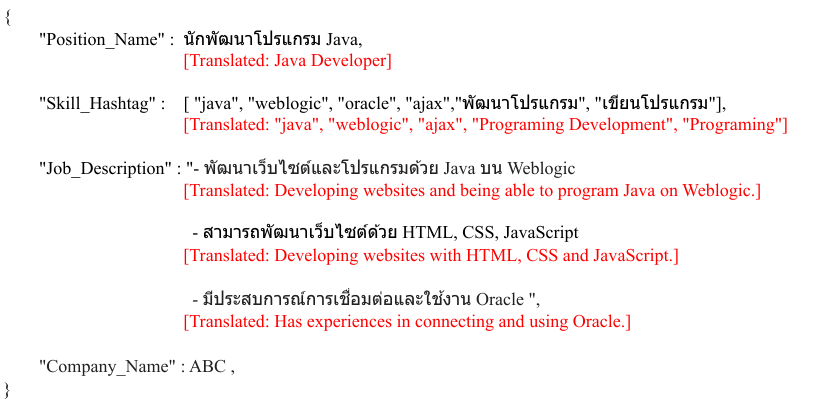}} 
    \caption{An example of job posting in the JTG-Jobposting dataset.}
    \label{fig:jtg-jobpost}
    
\end{figure*}

\begin{figure*}[ht]
    \centering
    \fbox{\includegraphics[width=0.8\textwidth]{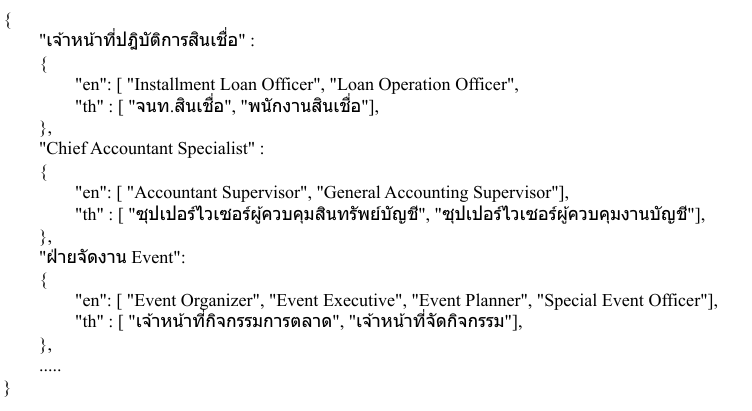}} 
    \caption{An example of synonym in the JTG-Synonym evaluation dataset.}
    \label{fig:jtg-syn}
    
\end{figure*}

\section{Detailed Dataset Description}
\label{sec:datasetdesc}


\subsection{JTG-Jobposting}
Dataset statistics of the JTG-Jobposting dataset are shown in Table \ref{tab:data-stats}. Overall, the dataset statistics are very similar to the Mycarrersfuture.sg dataset, though the former has a much higher number of distinct skills. This is because skill in the JTG-Jobposting dataset is more relaxed compared to Mycarrersfuture.sg where a predefined set of skills is available. Some data examples from JTG-Jobposting are shown in Figure \ref{fig:jtg-jobpost}. It consists of the following fields:
\begin{itemize}
    
\item \textbf{"Position\_Name"} : a job title of the job posting
\item \textbf{"Skill\_Hashtag"} : the skill tags assigned by the recruiter.
\item \textbf{"Job\_Description"}: a job description for the job posting.
\end{itemize}

\subsection{JTG-Synonym}
\label{apd:jtgsynonym}

The JTG-Synonym is a synonym list that includes different variants of the same job title. We split the list into validation and test of size 2,000 and 4,420, respectively. Thai and English titles were kept separated. Examples are shown in Figure \ref{fig:jtg-syn}. The statistics of the JTG-Synonym are shown in Table \ref{tab:stat-syn}.

\begin{table}[ht]
\centering
\resizebox{\columnwidth}{!}{%
\begin{tabular}{|l|c|ccc|cc|}
\hline
           & \multirow{2}{*}{\textbf{Total Query}} & \multicolumn{3}{c|}{\textbf{Query}}                                                       & \multicolumn{2}{c|}{\textbf{Candidate Pool}}          \\ \cline{3-7} 
           &                                       & \multicolumn{1}{c|}{\textbf{Thai}} & \multicolumn{1}{c|}{\textbf{English}} & \textbf{Code-Switching} & \multicolumn{1}{c|}{\textbf{Thai}} & \textbf{English} \\ \hline
Validation & 2,000                                 & \multicolumn{1}{c|}{1,044}         & \multicolumn{1}{c|}{928}              & 28           & \multicolumn{1}{c|}{7,762}         & 8,033            \\ \hline
Test       & 4,420                                 & \multicolumn{1}{c|}{2,261}         & \multicolumn{1}{c|}{2,103}            & 56           & \multicolumn{1}{c|}{16,905}        & 17,684           \\ \hline
\end{tabular}%
}
\caption{The statistics of query and candidate pool in validation and test set for JTG-Synonym. "Code-Switching" refers to the job title that contains both Thai and English.}
\label{tab:stat-syn}
\end{table}

\section{Extended Implementation Detail}

This subsection further describes our method and competing approaches. The model weight and inference code are available at \url{https://github.com/SLSCU/JD-agg-network}.

\label{apd:impdetail}

\subsection{Linear Probing}
\label{apd:linprobe}


 We applied linear probing by freezing the whole model except for the last feedforward layer. The performance of linear probing was evaluated on another set of JTG-Jobposting data to ensure no overlapping with the training data. The objective of the task is to predict the appropriate skills given the job titles. The dataset contains 6,861 training samples and 2,000 testing samples consisting of 157 classes (skills). Due to the sparsity of skill labeling, we evaluated the performance using top-10 accuracy. The experiment was conducted using job title embeddings from three sources: our skill-based method with recruiter annotation, JD-based method, and XLM-R without fine-tuning.
 

\subsection{Skill Extraction Model}
\label{apd:skillextraction}

The model used for skill extraction is mUSE \citep{yang-etal-2020-multilingual} followed by a 2-layer MLP with cross-entropy loss as the objective. During inference, we select the top 10 scores as candidates for representing the skill tags for each job posting. The model was trained on another separate set of the JTG-Jobposting dataset containing 12,240 samples, totaling 35,107 skills, that do not overlap with the original JTG-Jobposting dataset.


\subsection{Comparison against other methods}

Since the performance of the skill-based method of the JTG-Jobposting was not available, we reimplemented the baselines using the following configurations:


\begin{itemize}
\item \textbf{JobBERT}: We followed \citep{Decorte2021JobBERTUJ} by randomly selected five samples from a distribution defined by the frequency distribution of skills in the training corpus, raised to the power of 3/4 for training using the skip-gram technique. We used a batch size of 64 and a learning rate of 5e-6. 

\item \textbf{Doc2VecSkill}: We reimplemented this baseline by aggregating the skill set in each job title, and then Doc2vec was used to convert this set into auxiliary skill embeddings. The Doc2vec model was trained for 100 epochs with a dimension size of 768. Then, we matched the auxiliary skill embeddings with their embedded job titles using the cosine similarity loss. We used a batch of 64 and a learning rate of 3e-5.
\end{itemize}

\section{Qualitative Results}
\label{sec:attention_map}

\begin{figure*}[ht]
    \centering
    \includegraphics[width=0.8\textwidth]{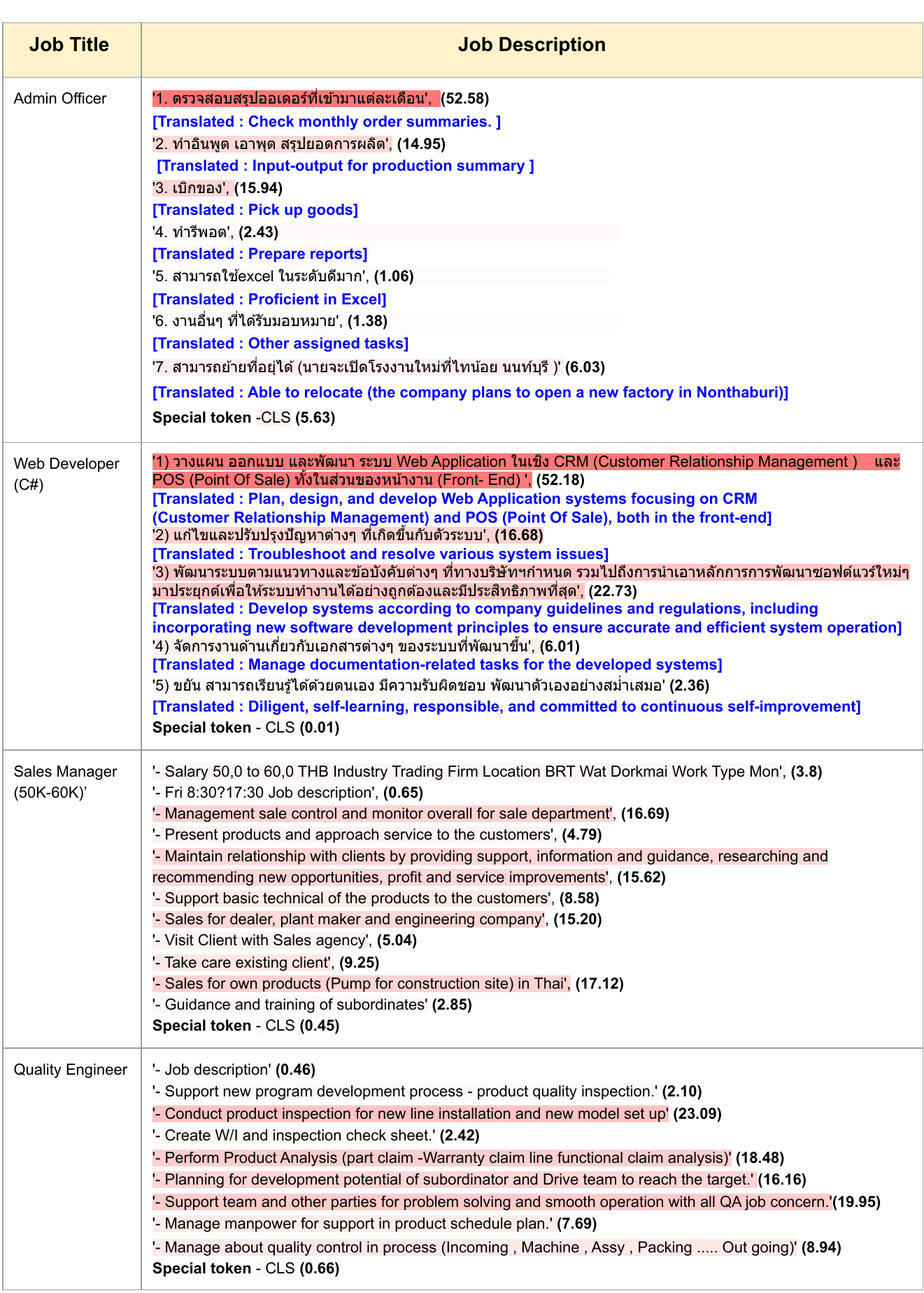} 
    \caption{Attention scores from the CLS token of sentences extracted from the first attention layer of the job description aggregator. Darker red means higher attention scores and vice versa.}
    \label{fig:heatmap}
    
\end{figure*}

Figure \ref{fig:heatmap} shows examples of attention maps in the first attention layer from the job description aggregator. It was found that the aggregator could correctly attend to sections with high importance and ignore the sentences unrelated to the job title. For example, in row 2 of Figure \ref{fig:heatmap}, the sentences "managing document of the developed software" and "Being studious, self-taught, responsible, and self-improvement" were mostly ignored while the sentence "Design and develop CRM Web Application ..." was strongly attended. The observation also held even when the first line was not the most informative sentence (example number 3 and 4).

    


\section{Design choice for our skill-based method}
\label{sec:appendixB}


We explored different design choices for combining multiple skills to train our skill-based method. These included averaging, maximizing the output of embedded skills, and concatenating the skill list. The results in Table \ref{tab:my-table} suggest that concatenation, our final choice, performed best.

\begin{table}[ht]
\centering
\resizebox{\columnwidth}{!}{%
\begin{tabular}{@{}lcc@{}}
\toprule
\multirow{2}{*}{}    & \textbf{JTG-Synonym}           & \textbf{Job Normalization}     \\ \cmidrule(l){2-3} 
                     & \textbf{R@10} $\uparrow$                          & \textbf{R@10}  $\uparrow$                         \\ \midrule
\textbf{Skill-Based} &                                &                                \\
Concat               & \textbf{60.62} ($\pm$0.32) & \textbf{56.45} ($\pm$0.18) \\
Max                  & 52.63 ($\pm$0.45) & 53.19 ($\pm$0.45) \\
Mean                 & 58.51 ($\pm$0.54) & 55.83 ($\pm$0.20) \\
\bottomrule
\end{tabular}%
}
\caption{The comparison of different design choices of the skill-based method.}
\label{tab:my-table}
\end{table}

\section{Design choice for the sentence representation aggregation}
\label{apd:aggeabl}
We explored different approaches for representing the sentence embedding from a sequence of tokens by comparing token aggregation using an average with directly using the [CLS] token. It was found that averaging the tokens in the sentence yielded marginal performance improvement over the usage of the [CLS] token.

\begin{table}[ht]
\centering
\begin{tabular}{lc}
\hline
\multirow{2}{*}{} & \textbf{Job Normalization} \\ \cline{2-2} 
                  & R@10  $\uparrow$           \\ \hline
Mean              & 57.09 ($\pm$0.39)                 \\
{[}CLS{]}         & 56.84 ($\pm$0.16)                
\end{tabular}%
\caption{JD-based (Document Level): Ablation of different method: "Mean" denotes averaging each tokens (words), "[CLS]" denotes using the [CLS] token.}
\label{tab:design-choice}
\end{table}

\end{document}